\documentclass[runningheads]{llncs}

\usepackage{eccv}

\usepackage{eccvabbrv}

\usepackage{graphicx}
\usepackage{booktabs}
\usepackage{booktabs}
\usepackage{multirow}
\usepackage{booktabs}
\usepackage{capt-of} % 提供 \captionof{table}{...}

\usepackage[accsupp]{axessibility}  % Improves PDF readability for those with disabilities.

\usepackage{hyperref}

\usepackage{orcidlink}

\begin{document}

\title{RPM-Distill: Physiology-guided Adaptive Cross-modal Distillation for Robust Remote Physiological Measurement} 

\titlerunning{RPM-Distill}

\author{Jiyao Wang\inst{1}\orcidlink{0000-0002-0743-0121} \and
Qingyong Hu\inst{2}\orcidlink{0000-0002-8719-7816} \and
Duoxun Tang\inst{4}\orcidlink{0009-0001-0463-8023} \and
Xiao Yang\inst{3}\orcidlink{0000-0001-6096-9903} \and
 Kaishun Wu\inst{3}\orcidlink{0000-0003-2216-0737} \and
 Jiangbo Yu\inst{1}\thanks{Corresponding author}\orcidlink{0000-0002-4525-4640}}

\authorrunning{J.~Wang et al.}

\institute{McGill University, Montreal, Canada \and
Hong Kong University of Science and Technology, Hong Kong, China\and
Hong Kong University of Science and Technology (Guangzhou), Guangzhou, China\and
Tsinghua University, Shenzhen, China\\
\email{\{jiangbo.yu\}@mcgill.ca}}

\maketitle

\begin{abstract}

Video-based remote physiological measurement (RPM) is highly accessible but remains fragile under varying illumination, skin tones, and motion. Radio frequency (RF) radar is largely invariant to illumination and appearance, providing complementary cardio-respiratory micro-motion cues; however, requiring radar at inference is often impractical due to its limited ubiquity and deployment overhead. We propose \textbf{RPM-Distill}, a physiology-guided cross-modal distillation framework that leverages synchronized radar only during training while retaining video-only inference. Our key observation is that although RGB and RF waveforms differ in sensing physics and time-domain morphology, they share similar latent periodic rhythm in the frequency domain. We thus distill physiology-structured spectral evidence to improve robustness, via losses that (i) anchor the fundamental peak, (ii) match the off-peak background distribution, and (iii) preserve spectral morphology and sharpness. To avoid negative transfer under sample-level teacher quality and alignment uncertainty, a spectral policy network predicts sample-level distillation gates and component weights from the student--teacher spectral relation map, learned with a meta bilevel objective on a small labeled validation split. Through extensive experiments in challenging conditions and cross-dataset settings, RPM-Distill brings 81\% MAE and 21\% correlation improvement over unimodal baselines. Code is at \url{https://github.com/WJULYW/RPM-Distill}.

\end{abstract}

\section{Introduction}
\label{sec:intro}

Remote physiological measurement (RPM) recovers vital signs, such as heart rate (HR) and blood volume pulse (BVP), without direct skin contact, enabling continuous monitoring in extensive applications \cite{wang2026drowsydg, wang2026mild}. Remote photoplethysmography (rPPG) is a representative RGB video-based RPM technique that extracts physiological signals from subtle, periodic appearance variations in facial or skin videos \cite{poh2010advancements,verkruysse2008remote}. 
Compared to contact-based acquisition, rPPG is non-invasive and highly accessible, yet it remains fundamentally fragile under unconstrained real-world conditions. Motion-induced non-rigid deformation and pose variations introduce strong temporal artifacts, while ambient light fluctuations and diverse skin tones significantly degrade the optical signal-to-noise ratio (SNR) \cite{nowara2021benefit,li2023learning,acharya2025reliability}. Improving the robustness of video-based RPM against these compounded interferences remains a central focus in RPM.

Radio-frequency (RF) radar provides a complementary sensing mechanism by measuring cardio-respiratory micro-displacements via phase modulation of reflected electromagnetic waves \cite{li2013review,adib2015smart}. It is generally invariant to illumination and appearance and can be privacy-friendly \cite{zhao2018rf}. These complementary properties have motivated video--radar fusion methods \cite{vilesov2022blending,ying2025fusionphys,liang2025spatial,ge2025evidential,choi2024fusion}, yet most existing solutions rely on multimodal processing at inference and require sufficiently large paired datasets for training, both of which are difficult to satisfy in practical deployments where cameras are ubiquitous, but radar is not.

\begin{figure*}[t]
    \centering
    \includegraphics[width=0.9\linewidth]{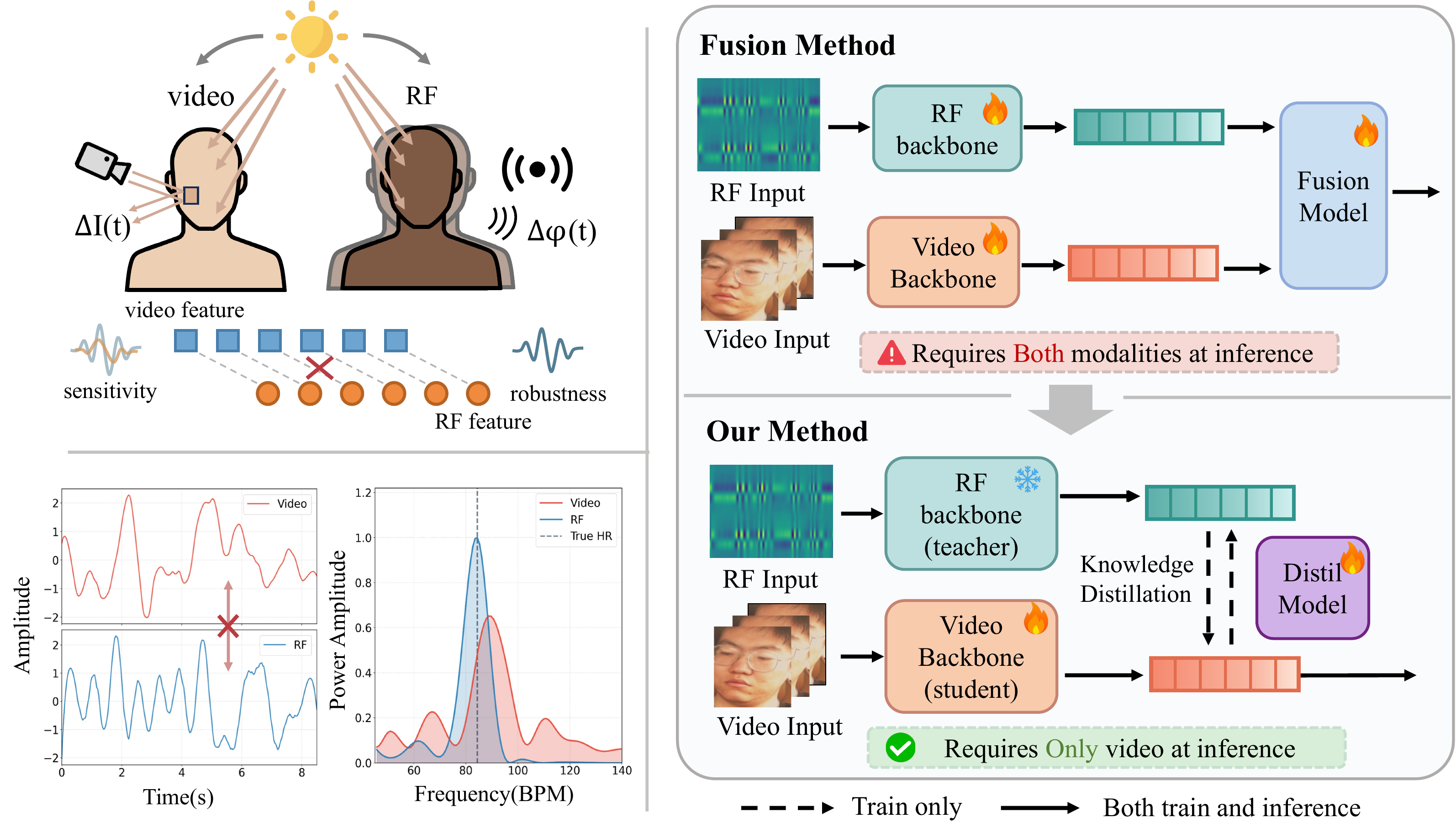}
    \vspace{-4mm}
        \caption{Motivation for our study. Left: rPPG measures pulsatile reflectance changes, whereas RF radar measures cardio-respiratory micro-displacements; their sensing physics yield mismatched time-domain morphology, making naive feature/time alignment prone to negative transfer. However, within the physiological band, both modalities are modulated by the same periodic rhythm, motivating the distillation on frequency spectra. Right: Instead of fusion that requires both sensors at test time, we distill training-time radar knowledge into a video-only student.}
    \label{fig:challenge}

\end{figure*}

This tension motivates a practical question: \textit{how can one exploit radar's complementary cues without committing to multi-sensor inference?} Cross-modal knowledge distillation (KD) \cite{huo2024c2kd,li2025mst} is a natural candidate, but it is particularly non-trivial for RPM due to physiology- and physics-induced modality mismatch. As summarized in Fig. \ref{fig:challenge}, first, RGB and RF observe the same underlying rhythm through fundamentally different transduction mechanisms, so their intermediate features and time-domain waveforms are not directly commensurate; feature-level alignment therefore suffers from modality/semantic mismatch \cite{baltruvsaitis2018multimodal}. Second, RPM distillation must be reliability-aware. Teacher quality and cross-modal alignment can vary across samples (e.g., transient corruption or synchronization shift), so indiscriminate output matching can propagate erroneous spectral evidence and cause negative transfer \cite{cho2019efficacy,wang2025physdrive}. Third, RPM targets are inherently periodic, and their failure modes are structured in the frequency domain. Under low SNR, video predictions often exhibit peak suppression/shift and spurious maxima, while nuisance factors elevate the off-peak noise floor and smear the local spectral shape \cite{niu2019rhythmnet,song2021pulsegan,nowara2021benefit,xi2020image}. These considerations suggest that effective cross-modal distillation for RPM should leverage the shared latent rhythm, and be sample-selective to avoid transferring unreliable teacher signals.

In this work, we address these challenges with \textbf{RPM-Distill}, a physiology-guided dynamic cross-modal distillation framework in a training-time privileged-information setting \cite{vapnik2009new}. We distill on band-limited frequency spectra where both modalities share the same latent periodicity, and impose a minimal yet complementary set of physiology-structured constraints that directly target dominant RPM failure modes: (i) fundamental peak anchoring for rhythm localization, (ii) off-peak/background matching for noise-floor suppression, and (iii) spectral morphology/sharpness consistency for stable rhythm evidence. To further handle sample-dependent teacher quality and alignment uncertainty, we introduce a spectral policy learner that predicts per-sample distillation gates and component weights from a three-channel student--teacher spectral relation map. Since optimal gating/weighting is unobserved and paired labeled data are limited, the policy is trained via a bilevel meta objective that is anchored by a small held-out validation split. Our contributions are summarized as:

\begin{itemize}
    \item To the best of our knowledge, we present the first cross-modal distillation framework for RPM, enabling video-only inference while leveraging RF teacher as the training-time privileged information. 
    \item We propose the physiology-guided spectral distillation objective with three components that target complementary spectral properties and mitigate modality-specific artifacts.
    \item We introduce an adaptive distillation policy network optimized via bilevel meta-learning. By predicting per-sample gates and weights from a three-channel spectral relation map, the policy improves both generalization and distillation selectivity under unreliability in teacher and synchronization.
    \item Extensive experiments on challenging conditions demonstrate consistent gains over both fusion, unimodal, and generic KD baselines, with improved generalization under diverse skin tone, motion, and dynamic lights.

\end{itemize}

\section{Related Works}
\noindent\textbf{Remote Physiological Measurement.} RPM has been primarily advanced by video-based methods that capture subtle facial color variations. Early approaches employed signal decomposition \cite{poh2010advancements, poh2010non,tu2016respiration} and color transformations \cite{wang2019discriminative, wang2016algorithmic}, yet remained constrained by strong priors. Deep learning subsequently enabled robust spatiotemporal modeling via supervised \cite{zou2025rhythmmamba, joshi2024factorizephys, yu2019remote, wang2026align,martinez2025beatformer,10903997} and unsupervised frameworks \cite{gideon2021way, speth2023non, sun2022contrast}. Nevertheless, they remain susceptible to illumination changes and motion artifacts \cite{li2014remote,wang2024condiff}. 

RF-based methods circumvent optical limitations by sensing cardiopulmonary-induced chest vibrations. Initial efforts used handcrafted decomposition \cite{li2008random, mercuri2018direct, tu2016respiration,alizadeh2019remote}; recent deep learning adaptations improved accuracy \cite{zheng2021more,khan2022contactless,hu2024contactless,hu2025mmtremor}. However, RF-based models still suffer from electromagnetic interference and occlusion-induced attenuation.

Recognizing their complementary nature, early works explored Video-RF fusion \cite{vilesov2022blending, choi2024fusion}, yet struggled with the modality gap and asynchronous sensor degradation. Recent advances propose adaptive strategies: dynamically weights modalities by uncertainty and noise \cite{ge2025evidential}; integration with cross-modal temporal dynamics \cite{ying2025fusionphys}; and adapter-based alignment mitigates spatial-temporal mismatches \cite{liang2025spatial}. Nevertheless, some feature-level fusion methods \cite{liang2025spatial} account for alignment and impose dependence on particular backbone architectures of both modalities, which may impair the transferability. Besides, all existing fusion approaches still require multimodal input in inference, which hinders deployment.

\noindent\textbf{Privileged-Information Learning and Adaptive Distillation.} Learning Using Privileged Information (LUPI) formalizes the training–deployment mismatch where extra cues are available only during training, improving generalization via additional supervision signals \cite{vapnik2009new}. Generalized distillation unifies LUPI and KD by treating the privileged modality as a teacher that provides soft targets to a student with limited inputs at deployment \cite{lopez2015unifying}. KD typically transfers "dark knowledge" through softened outputs \cite{hinton2015distilling} and has been extended to intermediate features and attention maps \cite{zagoruyko2017paying,romero2015fitnetshintsdeepnets,zhang2018deep}.

Cross-modal KD instantiates when teacher and student use different modalities. Early work showed that a stronger modality can supervise a weaker one via cross-modal distillation for supervision transfer \cite{gupta2016cross,hoffman2016learning}. In video understanding, auxiliary streams (e.g., depth/optical flow) have been used as training-time privileged information, while keeping RGB-only testing \cite{garcia2018modality}. Recent cross-modal KD methods go beyond static objectives to address modality gaps and sample-dependent transfer quality \cite{huo2024c2kd,li2025mst,guomultimodal}, particularly in VLM pretraining and efficient adaptation \cite{kim2025cosmos,feng2025align}. Some works also attempt to distill knowledge between RF and video \cite{pokhrel2023deakin,shan2024identitykd} for human/object detection, while cross-modal KD towards RPM remain under explored. The two modalities share the same latent rhythm but arise from different sensing physics and exhibit structured, sample-dependent spectral corruption in RPM. Naive global matching might propagate modality-specific artifacts, which motivates physiology-aware distillation on spectral components.

%However, most cross-modal KD studies still focus on semantic recognition \cite{zhao2024crkd,zhou2023unidistill}.

\section{Methodology}
\label{sec:method}

\subsection{Preliminaries}
\label{sec:prelim}

% \textbf{Motivation.} rPPG extracts the BVP from facial videos, yet it remains highly susceptible to illumination variations and motion artifacts. RF radar senses cardio-respiratory micro-displacements with illumination invariance, providing complementary physiological cues often obscured in the optical domain. However, existing fusion approaches typically require multimodal inputs during inference. To transcend this limitation, we formulate a cross-modal distillation framework, RPM-Distill, where RF radar serves as training-time privileged information. The overview of our proposal is shown in Fig. \ref{fig:model}.

\begin{figure*}[t]
    \centering
    \includegraphics[width=0.9\linewidth]{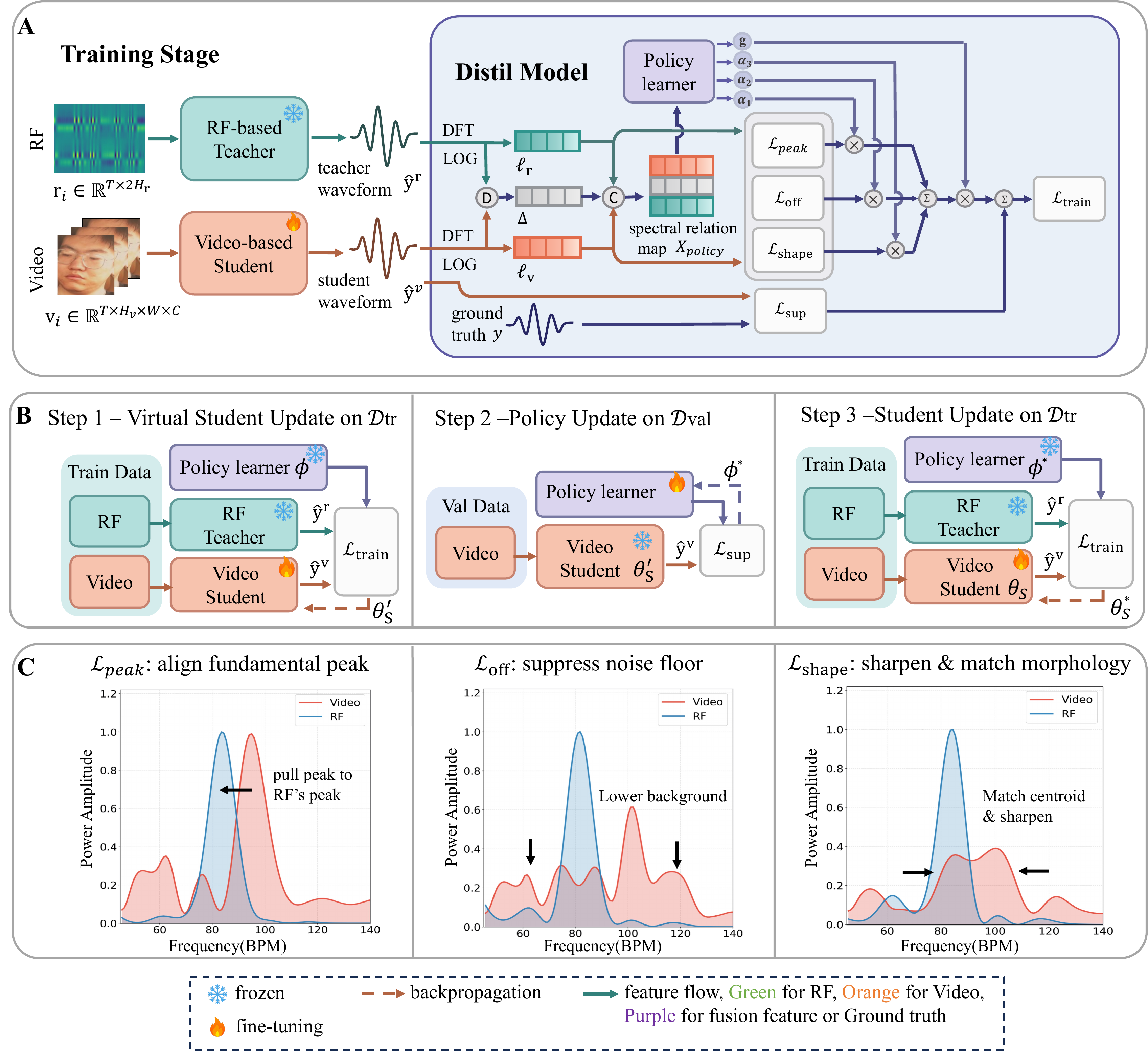}
    \vspace{-2mm}
    \caption{The overview of RPM-Distill. (A) In training, a fixed radar teacher provides guidance to a video student via distillation with three components. A spectral policy network predicts a per-sample gate and component weights from the student--teacher spectral relation map, and is optimized by bilevel meta-learning. (B) The process of bilevel meta optimization. A virtual student update is evaluated on a labeled validation split to update the policy, after which the student is updated using the refined policy. (C) The illustration of our three distillation components in the frequency domain.}
    \label{fig:model}

\end{figure*}

%\noindent\textbf{Problem Formulation.}
We consider a multimodal dataset $\mathcal{D}_{tr} = \{(\mathbf{v}_i, \mathbf{r}_i)\}_{i=1}^N$ during training. $\mathbf{v}_i \in \mathbb{R}^{T \times H_v \times W \times C}$ is a facial image sequence, where $T, C$ is the number of frames and channels, and $H_v$ and $W$ are height and width, respectively. For RF data, the raw signal is preprocessed into a range--time representation and decomposed into in-phase and quadrature components, resulting in a feature tensor $\mathbf{r}_i \in \mathbb{R}^{T \times 2H_r}$ (see Supplementary for details). Ground-truth physiological labels $\mathbf{y}$ may be only available for a small-scale training data.

Our goal is to learn a video-based student network $S(\cdot; \theta_S)$ that predicts a physiological waveform $\hat{\mathbf{y}}^v = S(\mathbf{v}_i; \theta_S)$ using only RGB videos at inference. During training, we leverage a pre-trained radar teacher $R(\cdot; \theta_R)$ to provide auxiliary guidance $\hat{\mathbf{y}}^r = R(\mathbf{r}_i; \theta_R)$. Depending on label availability, we optimize the student with either distillation-only training or joint supervised-and-distilled training. This is summarized by the following label-optional objective:
\begin{align}
\theta_S^* &= \arg\min_{\theta_S} \mathbb{E}_{(\mathbf{v}, \mathbf{r}) \sim \mathcal{D}_{tr}} \Big[ \mathcal{L}_{\text{train}}\!\big(\hat{\mathbf{y}}^v,\mathbf{y}\big) \Big], \\ \nonumber
\mathcal{L}_{\text{train}}\!\big(\hat{\mathbf{y}}^v,\hat{\mathbf{y}}^r,\mathbf{y}\big) &= \mathbb{I}_{\{\mathbf{y}\}} \mathcal{L}_{\text{sup}}\!\big(\hat{\mathbf{y}}^v,\mathbf{y}\big) + \lambda_{\text{distill}} \mathcal{L}_{\text{distill}}\!\big(\hat{\mathbf{y}}^v,\hat{\mathbf{y}}^r;\phi\big).
\label{eq:distill_objective}
\end{align}

$\mathbb{I}_{\{\mathbf{y}\}}$ is an indicator function that equals 1 if the ground-truth physiological label $\mathbf{y}$ is available for the current sample, and 0 otherwise. $\mathcal{L}_{\text{distill}}$ is governed by a meta-policy $\phi$. To prevent the policy from propagating modality-specific noise, a separate labeled validation set $\mathcal{D}_{val}$ is reserved exclusively to meta-optimize $\phi$, anchoring the learned policy to supervised physiological recovery. The overview of our proposal is shown in Fig. \ref{fig:model}.

\subsection{Physiology-Structured Spectral Distillation}
\label{sec:distill_obj}

Physiological signals are fundamentally periodic, yet standard cross-modal distillation is hard to directly apply to RPM. RGB and RF follow different sensing physics, their time-domain waveforms and intermediate features are not commensurate, and indiscriminate alignment in feature or time domains can propagate modality-specific artifacts and amplify errors when the teacher is locally unreliable. We instead exploit the fact that, despite distinct morphology, both modalities encode the same latent cardiac rhythm in the frequency domain. This shared spectral structure provides a more stable interface for supervision and reduces sensitivity to modality-dependent distortions (e.g., radar multipath or optical shading). Moreover, the dominant RPM failure modes manifest in a small set of recurring spectral patterns, including peak mis-localization, off-peak background elevation, and local spectral smearing. We therefore design a new distillation objective by decomposing supervision into three complementary spectral components that explicitly target these modes.

Specifically, for a predicted waveform $\hat{\mathbf{y}}$ of each modality $m\in\{v,r\}$, we compute its band-limited power spectrum $\mathbf{P}^m\in\mathbb{R}^{K}$ and log-spectrum $\boldsymbol{\ell}^m\in\mathbb{R}^{K}$ that is $z$-score normalized \emph{per sample} over frequency bins within the physiological band $\mathcal{B}$, where $k\in\mathcal{B}$ indexes discrete frequency bins within the physiological band, and $K=|\mathcal{B}|$ denotes the number of bins (details in Supplementary):

\begin{equation}
\mathbf{P}^m[k]=\left|\mathcal{F}(\hat{\mathbf{y}}^m)[k]\right|^2,\quad
\boldsymbol{\ell}^m[k]=\log(\mathbf{P}^m[k]+\epsilon),\quad k\in\mathcal{B}.
\end{equation}

$\mathcal{F}$ denotes the Discrete Fourier Transform (DFT) and $\epsilon$ is a small constant for numerical stability. This operation yields $(\mathbf{P}^v,\boldsymbol{\ell}^v)$ for the student and $(\mathbf{P}^r,\boldsymbol{\ell}^r)$ for the teacher. Their element-wise absolute discrepancy is $\Delta = |\boldsymbol{\ell}^v - \boldsymbol{\ell}^r| \in \mathbb{R}^{K}$.

\noindent\textbf{Component Decomposition.} To construct differentiable masking mechanisms, we define the soft spectral centroid \cite{sun2018integral} over the normalized distribution $\mathbf{s}^m=\mathrm{softmax}(\boldsymbol{\ell}^m) \in \mathbb{R}^{K}$ as $\hat{k}^m \triangleq \sum_{k\in\mathcal{B}} k\cdot \mathbf{s}^m[k]$. To localize supervision around the teacher-provided frequency anchor, we subsequently generate a Gaussian soft window mask centered at the teacher peak $\hat{k}^r$.

\begin{equation}
\mathbf{m}_{\mathrm{pk}}[k]=\exp\!\left(-\frac{(k-\hat{k}^r)^2}{2\sigma^2}\right),
\end{equation}

where the $\sigma\approx0.01K$ controls the window bandwidth. The remaining-band mask is defined as
$\mathbf{m}_{\mathrm{oth}}=\max(\mathbf{0}, \mathbf{1}-\mathbf{m}_{\mathrm{pk}})$. Utilizing these spatial-frequency masks, we establish three physiological constraints:

\textit{1) Fundamental Peak Alignment ($\mathcal{L}_{\text{peak}}$).} For cardiac RPM, the dominant spectral mode within the physiological band encodes the fundamental rhythm and directly determines HR \cite{speth2023non}. In practice, low illumination and skin-tone-dependent attenuation reduce rPPG SNR, causing the student spectrum to exhibit peak shifts or spurious dominant peaks away from the true rhythm \cite{nowara2020meta,nowara2021benefit}. By contrast, when the radar teacher is reliable, its fundamental peak provides a stable frequency anchor even when the optical waveform is temporally distorted. Motivated by this, we enforce agreement specifically within a narrow neighborhood of the teacher peak, so that distillation corrects peak mis-localization without requiring time-domain waveform comparability: 

\begin{equation}
    \mathcal{L}_{\text{peak}}= \frac{\|\Delta \odot \mathbf{m}_{\mathrm{pk}}\|_1}{\|\mathbf{m}_{\mathrm{pk}}\|_1+\epsilon}.
\end{equation}

\textit{2) Background Noise Suppression ($\mathcal{L}_{\text{off}}$).} Another pervasive failure mode of video-based RPM is broadband spectral pollution. Motion, facial expressions, dynamic lighting, and imperfect ROI tracking inject non-physiological dynamics that spread energy across the band and elevate the off-peak noise floor \cite{li2023learning,wang2016algorithmic}. Such distributed interference is harmful even when the dominant peak is approximately correct, because it destabilizes peak selection and reduces the effective spectral contrast between the rhythm and background. Empirically, when the teacher is reliable, radar spectra tend to concentrate energy more strongly around the cardiac component while exhibiting a lower background level. We therefore match the student and teacher outside the peak window, encouraging the student to suppress non-peak energy:

\begin{equation}
    \mathcal{L}_{\text{off}}= \frac{\|\Delta \odot \mathbf{m}_{\mathrm{oth}}\|_1}{\|\mathbf{m}_{\mathrm{oth}}\|_1+\epsilon}.
\end{equation}

\textit{3) Spectral Morphology Consistency ($\mathcal{L}_{\text{shape}}$).} Peak frequency alone is insufficient for robust RPM, because nuisance factors often preserve an approximate center frequency while altering the spectral morphology. In particular, motion and illumination transients can broaden the cardiac component in the student spectrum (peak smearing), leading to ambiguous maxima and reduced temporal consistency, whereas a reliable teacher typically exhibits a more concentrated spectral mass \cite{wang2016algorithmic}. This motivates a morphology-level constraint that captures two complementary properties: (i) peak localization, which penalizes systematic shifts of the spectral centroid, and (ii) spectral sharpness \cite{elgendi2024optimal}, which penalizes overly diffuse distributions and reflects the confidence of the inferred rhythm. Using the normalized spectral distribution $\mathbf{s}$, we quantify sharpness via entropy and combine both terms with Smooth L1 $\mathcal{L}_{l1}$:

\begin{align}
H(\mathbf{s})&=-\frac{1}{\log K}\sum_{k\in\mathcal{B}} \mathbf{s}[k]\log(\mathbf{s}[k]+\epsilon), \nonumber\\
\mathcal{L}_{\text{shape}}&= \mathcal{L}_{l1}(\hat{k}^v,\hat{k}^r)+ \mathcal{L}_{l1}\big(H(\mathbf{s}^v),H(\mathbf{s}^r)\big).
\end{align}

\noindent\textbf{Unified Distillation Objective.} The final objective is a policy-weighted summation. A global distillation gate $g\in(0,1)$ and component weights $\boldsymbol{\alpha}\in\Delta^{|\mathcal{J}|-1}$ are dynamically predicted by the meta-policy $\phi$. We define $\mathcal{J}=\{\text{peak},\text{off},\text{shape}\}$:

\begin{equation}
\mathcal{L}_{\text{distill}}(\hat{\mathbf{y}}^v,\hat{\mathbf{y}}^r;\phi)
= g\cdot\sum_{j\in\mathcal{J}}\boldsymbol{\alpha}_j\,\mathcal{L}_j.
\end{equation}

\subsection{Data-Driven Spectral Policy Network}
\label{sec:policy}

As the work \cite{ge2025evidential} reveals, enforcing alignment indiscriminately is highly susceptible to negative transfer. Towards the optimal distillation strategy, we design a data-driven spectral policy network $\Phi(\cdot; \phi)$ to evaluate inter-modal dependencies and adaptively modulate the distillation process from the spectral topology.

To provide the policy with a comprehensive physiological context, we construct a localized spectral relation map directly from the frequency domain. Specifically, for a given video-radar sample pair, we utilized the log-PSD vectors of both the student prediction $\boldsymbol{\ell}^v$ and the teacher prediction $\boldsymbol{\ell}^r$. We stack them alongside their absolute discrepancy $\Delta$ to form a three-channel spectral relation map $\mathbf{X}_{\text{policy}} =[\boldsymbol{\ell}^v, \boldsymbol{\ell}^r, \Delta]^T , \in \mathbb{R}^{3 \times K}$. This tensor explicitly preserves the localized correlation structure across the frequency dimension $K$.

Subsequently, $\mathbf{X}_{\text{policy}}$ passes through our spectral policy learner. The encoder consists of a 1D convolutional stem and successive residual blocks equipped with group normalization to extract multi-scale spectral representations. Then, inspired by \cite{gengattention,joshi2024ibvp}, we introduce a matrix-decomposition decoder, which learns basis and coefficient attentions over the frequency axis to extract low-rank quality tokens. These tokens are then fused with globally pooled features and bifurcated through linear projection heads to derive the optimal modulation factors:

\begin{equation}
    g, \boldsymbol{\alpha} = \Phi(\mathbf{X}_{\text{policy}}; \phi).
\end{equation}

Here, $g$ is bounded by a Sigmoid activation, indicating the distillation's instantaneous fidelity. Concurrently, the component weighting vector $\boldsymbol{\alpha}$ is normalized via a Softmax operation to reside on the probability simplex $\Delta^{|\mathcal{J}|-1}$.

\subsection{Bilevel Meta-Optimization}
\label{sec:meta}

Optimizing the policy network $\Phi(\cdot; \phi)$ introduces a challenge: we lack explicit ground truth labels for the optimal gate $g$ and weights $\boldsymbol{\alpha}$. Manually tuning these hyperparameters is intractable given the dynamic, sample-dependent nature of unreliable distillation. To resolve this, we formulate the learning of $\phi$ as a bilevel optimization problem \cite{ji2021bilevel,sinha2017review}. The core insight is to treat the distillation policy as a learnable inductive bias. 

Let $\theta_S$ denote the student parameters and $\phi$ the policy parameters. We aim to find the optimal $\phi^*$ that minimizes the student's supervised loss $\mathcal{L}_{\text{sup}}$ on a held-out validation set $\mathcal{D}_{val}$, subject to the constraint that $\theta_S$ is optimized on the training set $\mathcal{D}_{tr}$ using $\mathcal{L}_{\text{distill}}$. In this work, the supervised loss $\mathcal{L}_{\text{sup}}$ is instantiated as a combination of negative Pearson correlation loss \cite{yu2019remote} and SNR loss \cite{vspetlik2018visual} to rigorously quantify the physiological waveform fidelity.

Then, since directly solving this nested optimization is computationally prohibitive, we adopt an online approximation strategy that interleaves the updates of the student and the policy. First, we simulate the student model's change trajectory when updated under the current policy $\phi$. Using $\mathcal{D}_{tr}$, we perform a virtual gradient descent step to obtain temporary student parameters $\theta'_S$, where $\eta$ is the learning rate:

\begin{equation}
    \theta'_S \leftarrow \theta_S - \eta \nabla_{\theta_S} \mathcal{L}_{\text{train}}(\theta_S,\phi;\mathcal{D}_{tr}).
\end{equation}

We then evaluate the supervised loss of the virtual student $\theta'_S(\phi)$ on $\mathcal{D}_{val}$ and update $\phi$ to favor distillation decisions that improve supervised generalization. Among Eq. (\ref{eq10}), $\beta$ is the meta-learning rate. This critical step explicitly aligns the distillation incentive with the student's empirical performance.

\begin{equation}\label{eq10}
    \phi^* \leftarrow \phi - \beta \nabla_{\phi} \mathcal{L}_{\text{sup}}(\theta'_S(\phi); \mathcal{D}_{val}).
\end{equation}

Finally, with the optimized policy $\phi^*$, we perform the actual update on the student network $\theta_S$ using $\mathcal{D}_{tr}$ to obtain the final $\theta_S^*$ (Eq. (\ref{eq11})). By coupling the spectral constraints (Sec.~\ref{sec:distill_obj}) with this meta-optimization, RPM-Distill establishes a closed-loop system where physiological priors define the solution space, while the data-driven policy navigates the optimization trajectory.

\begin{equation}\label{eq11}
    \theta^*_S \leftarrow \theta_S - \eta \nabla_{\theta_S} \mathcal{L}_{\text{train}}(\theta_S,\phi^*;\mathcal{D}_{tr}).
\end{equation}

\section{Experiments}

\subsection{Datasets and Baselines}
We conducted experiments on four datasets: \textbf{EquiPleth} \cite{vilesov2022blending} contains 91 participants' RGB video and radar data with three skin tones and well modality synchronization. \textbf{PhysDrive} \cite{wang2025physdrive} was collected in a naturalistic driving scenario with 48 subjects and comprises RGB video and radar data under varying driver motions, dynamic natural lighting. Its synchronization was on the second level. \textbf{PURE} \cite{stricker2014non} consists of RGB video recordings from 10 subjects under different activities. \textbf{MMPD} \cite{tang2023mmpd} consists of RGB videos from 33 participants performing four distinct activities and exercise scenarios.

We choose six RGB video-based rPPG baselines, including two traditional methods (CHROM \cite{de2013robust}, POS \cite{wang2016algorithmic}), and four deep-based methods (PhysNet \cite{yu2019remote}, PhysFormer \cite{yu2022physformer}, FactorizePhys \cite{joshi2024factorizephys}, BeatFormer \cite{martinez2025beatformer}). Besides, for RF-based methods, following \cite{wang2025physdrive}, we adopt one traditional FFT-based method \cite{alizadeh2019remote}, and four deep-based baelines (IQ-MVED \cite{zheng2021more}, VitaNet \cite{khan2022contactless}, mmFormer \cite{hu2024contactless}, and the RF predictor in \cite{vilesov2022blending}). For the RGB-RF multi-modal fusion baselines, we choose Vilesov et al. \cite{vilesov2022blending}, Fusion-Vital \cite{choi2024fusion}, FusionPhys \cite{ying2025fusionphys}, SATM \cite{liang2025spatial}, and EvidentialFusion \cite{ge2025evidential}. Lastly, we compare our proposal with four representative KD baselines, including feature-based FitNets \cite{romero2015fitnetshintsdeepnets}, DML with mutual information \cite{zhang2018deep}, C$^2$KD \cite{huo2024c2kd} and MST-Distill \cite{li2025mst} for cross-modal KD. For fair comparison, we disable the multimodal specialized teachers in MST-Distill.

\subsection{Implementation Details}
% The proposed method is implemented in PyTorch and trained on an NVIDIA RTX3090 GPU. For the input, we set the sampling rate to $30$\,Hz and restrict the physiological band $\mathcal{B}$ to $[45,180]$ bpm for all spectral operations and HR estimation. Each training video/rf sample is formed as a clip of length $T=256$ frames with a stride of 30 frames. We use the MTCNN \cite{zhang2016joint} to crop the face region and resize each frame
% to $\mathbb{R}^{72\times 72\times 3}$ for video. For RF, the window size of the maximum power occupancy of the range matrix $H_r=5$. The student $S(\cdot;\theta_S)$ is trained with Adam (weight decay $10^{-2}$) for 5 epochs, using a learning rate of $5\times 10^{-5}$ and batch size 5. The radar teacher $R(\cdot;\theta_R)$ is initialized from a supervised pretrained checkpoint on the labeled samples in the training set, kept frozen, and only used for producing $\hat{\mathbf{y}}^{r}$. According to the experiment results, we instantiate the $S(\cdot;\theta_S)$ with FactorizePhys, and $R(\cdot;\theta_R)$ as same in \cite{vilesov2022blending}. For the policy learner $\Phi(\cdot;\phi)$, we use the 1D ConvNet with base width 32, depth 3, dropout 0.1, and a matrix-decomposition decoder with rank 8 and hidden size 32 (details are in Supplementary). The policy outputs $g$ and $\boldsymbol{\alpha}$ and is optimized by Adam with meta learning rate $\beta=10^{-4}$. Bilevel updates are performed every 10 student steps, using a single differentiable virtual step with $\eta$ set equal to the student learning rate. We set $\lambda_{\text{distill}}=1.0$. 
The proposed method is implemented in PyTorch and trained on an NVIDIA A800 GPU. We set the sampling rate to $30$\,Hz and restrict the physiological band $\mathcal{B}$ to $[45,180]$\,bpm for all spectral operations and HR estimation. Each training sample is a clip of length $T=256$ with a stride of 30 frames. We use MTCNN \cite{zhang2016joint} for face cropping and resize frames to $72\times72$. We train the student with Adam ($\eta=5\times10^{-5}$, weight decay $10^{-2}$) for 5 epochs, and keep a supervised-pretrained radar teacher frozen for producing $\hat{\mathbf{y}}^{r}$. Unless otherwise stated, we instantiate the student with FactorizePhys and the teacher with the RF predictor in \cite{vilesov2022blending}. The spectral policy network is optimized with Adam ($\beta=10^{-4}$), with bilevel updates every 10 student steps. Details are in the Supplementary.

All models aim to recover the waveform. We compute HR for each signal clip using band-pass filtering, detrending, and frequency-domain peak detection. To evaluate the predicted HR error, we report the average standard deviation of the error (STD), mean absolute error (MAE), root mean square error (RMSE), and Pearson’s correlation coefficient (r) of five-time experiments with different random seeds. STD, MAE, and RMSE are reported in beats per minute (bpm).

\subsection{Experimental Results}

\begin{table*}[t]
\centering
\caption{Cross-dataset performance comparison on PhysDrive and EquiPleth dataset. In this and following tables, best results are in \textbf{bold} and second best are \underline{underlined}. Statistically significant ($p$<.005) improvements are marked '$^*$' using paired $t$-tests.}
\vspace{-2mm}
\label{tab:cross}
\setlength{\tabcolsep}{3pt}
\renewcommand{\arraystretch}{0.9}
\scriptsize
\begin{tabular}{llcccccccc}
\toprule
\multirow{2}{*}{Modality} & \multirow{2}{*}{Method} & \multicolumn{4}{c}{PhysDrive \cite{wang2025physdrive}} & \multicolumn{4}{c}{EquiPleth \cite{vilesov2022blending}} \\
\cmidrule(lr){3-6}\cmidrule(lr){7-10}
 &  & STD$\downarrow$ & MAE$\downarrow$ & RMSE$\downarrow$ & $r\uparrow$ & STD$\downarrow$ & MAE$\downarrow$ & RMSE$\downarrow$ & $r\uparrow$ \\
\midrule
\multirow{6}{*}{Video} & CHROM \cite{de2013robust} & 17.17 & 16.72 & 23.97 & 0.19 & 10.77 & 7.31 & 17.42 & 0.54 \\
 & POS \cite{wang2016algorithmic} & 17.33 & 17.38 & 24.38 & 0.20 & 11.51 & 4.64 & 15.07 & 0.68 \\
 & PhysNet \cite{yu2019remote} & \underline{12.73} & 18.45 & 25.45 & 0.15 & 11.10 & 7.93 & 13.64 & 0.55 \\
 & PhysFormer \cite{yu2022physformer} & 17.10 & 18.87 & 21.36 & 0.16 & 12.56 & 9.47 & 13.20 & 0.34 \\
 & FactorizePhys \cite{joshi2024factorizephys} & 14.74 & 15.67 & 20.55 & 0.19 & 11.83 & 8.21 & 14.40 & 0.78 \\
 & BeatFormer \cite{martinez2025beatformer} & 16.10 & 17.16 & 21.80 & 0.17 & 13.82 & 9.08 & 13.98 & 0.62 \\
\midrule
\multirow{5}{*}{RF} & FFT-based \cite{alizadeh2019remote} & 18.21 & 20.60 & 24.29 & 0.10 & 8.77 & 12.82 & 15.53 & 0.32 \\
 & IQ-MVED \cite{zheng2021more} & 16.61 & 18.61 & 20.45 & 0.12 & 8.60 & 13.06 & 15.64 & 0.34 \\
 & VitaNet \cite{khan2022contactless} & 18.31 & 16.31 & 21.07 & 0.16 & 8.36 & 11.88 & 14.52 & 0.45 \\
 & mmFormer \cite{hu2024contactless} & 15.38 & 15.38 & 27.03 & 0.15 & 12.41 & 13.66 & 13.85 & 0.55 \\
 & Vilesov et al. \cite{vilesov2022blending} & 13.60 & 15.21 & \underline{20.42} & 0.21 & 7.79 & 8.46 & 9.44 & 0.78 \\
\midrule
\multirow{4}{*}{Fusion} & Vilesov et al. \cite{vilesov2022blending} & 19.34 & 17.89 & 22.93 & 0.21 & 6.45 & 5.82 & 9.59 & 0.85 \\
 & FusionPhys \cite{ying2025fusionphys} & 15.91 & 15.35 & 22.48 & \textbf{0.23}$^{*}$ & \underline{5.24} & 4.88 & 6.48 & 0.89 \\
 & SATM \cite{liang2025spatial} & 15.22  & \textbf{13.79}$^{*}$  & 21.99  & \textbf{0.23}$^{*}$  & 5.75  & 3.92  & 6.62  & 0.88  \\
 & EvidentialFusion \cite{ge2025evidential} & 15.21 & 16.12 & 22.42 & 0.19 & 7.06 & 7.35 & \underline{6.04} & 0.85 \\
\midrule
\multirow{5}{*}{\shortstack{Cross-modal\\KD}} & FitNets \cite{romero2015fitnetshintsdeepnets} & 15.26 & 20.20 & 28.31 & 0.09 & 6.60 & 3.09 & 7.23 & 0.85 \\
 & DML \cite{zhang2018deep} & 15.25 & 17.96 & 23.58 & 0.13 & 5.44 & 2.57 & 7.98 & 0.87 \\
 & C$^2$KD \cite{huo2024c2kd} & 13.21 & 18.25 & 25.37 & 0.12 & 6.92 & \underline{2.45} & 6.88 & \underline{0.90} \\
 & MST-Distill \cite{li2025mst} & 14.69 & 16.55 & 23.47 & 0.17 & 5.58 & 2.99 & 7.31 & 0.87 \\
 & \textbf{Ours} & \textbf{12.67}$^{*}$ & \underline{14.15} & \textbf{20.24}$^{*}$ & \underline{0.22} & \textbf{4.14}$^{*}$ & \textbf{1.57}$^{*}$ & \textbf{4.64}$^{*}$ & \textbf{0.94}$^{*}$ \\
\bottomrule
\end{tabular}

\end{table*}

\textbf{Main results.} We first evaluate RPM-Distill under the cross-dataset protocol, where we train and validate on one dataset, then test and report on the other one. We split the train and validation set with 4:1 in subject level with random seeds. In this experiment, we optimize the student with both the supervised loss and the proposed distillation loss on the full training set. As shown in Table~\ref{tab:cross}, video-only rPPG baselines suffer substantial performance degradation under domain shift. Radar-only methods also exhibit limited transferability, suggesting that RF spectra still may provide unreliable supervision across domains.

Previous KD baselines mitigate the modality gap by feature/output transfer from RF to RGB, yet their improvements remain bounded, particularly when trained on PhysDrive, where dynamic lighting, driving-induced motion, and worse synchronization amplify the optical failure modes. In contrast, RPM-Distill achieves the best overall error metrics on both datasets and maintains strong correlation, even compared to multimodal fusion approaches, which utilize the multimodal data in inference. Notably, it reduces the MAE/RMSE on EquiPleth to 1.57/4.64, outperforming both fusion and KD baselines. These results support our key design: when training data is with more noise either inter-modal SNR or cross-modal alignment, fusion is exposed to RF corruption and misalignment; while our selective distillation on shared rhythm-relevant regions can effectively avoid negative transfer.

\begin{table*}[t]
\centering
\caption{Cross-dataset comparison between the best RGB-based baseline and our method under different conditions and limited labeled data. In this table, `E.M.\&D.' means `early morning and dusk', and `C.\&R.' is `cloudy and rainy day'.}
\vspace{-2mm}
\label{tab:ssl}
\setlength{\tabcolsep}{3pt}
\renewcommand{\arraystretch}{0.95}
\scriptsize

\resizebox{\linewidth}{!}{%
\begin{tabular}{lcccccccccccc}
\toprule
\multirow{2}{*}{Method} &
\multicolumn{2}{c}{E.M.\&D.} &
\multicolumn{2}{c}{Noon} &
\multicolumn{2}{c}{Night} &
\multicolumn{2}{c}{R.\&C.} &
\multicolumn{2}{c}{Stationary} &
\multicolumn{2}{c}{Talking} \\
\cmidrule(lr){2-3}\cmidrule(lr){4-5}\cmidrule(lr){6-7}\cmidrule(lr){8-9}\cmidrule(lr){10-11}\cmidrule(lr){12-13}
& MAE$\downarrow$ & $r\uparrow$ &
  MAE$\downarrow$ & $r\uparrow$ &
  MAE$\downarrow$ & $r\uparrow$ &
  MAE$\downarrow$ & $r\uparrow$ &
  MAE$\downarrow$ & $r\uparrow$ &
  MAE$\downarrow$ & $r\uparrow$ \\
\midrule
FactorizePhys \cite{joshi2024factorizephys} & 15.32 & 0.19 & 13.28 & 0.23 & 20.51 & 0.01 & 13.78 & 0.18 & \underline{13.55} & 0.20 & 18.31 & 0.05 \\
% \textbf{Ours} 0\%      & 13.57 & 0.23 & 14.71 & 0.26 & 19.82 & 0.05 & 14.37 & 0.15 & 16.68 & 0.13 & 17.93 & 0.09 \\
 \textbf{Ours} 20\%  & 12.42 & \underline{0.28} & 14.01 & 0.27 & 18.33 & \underline{0.06} & 14.38 & 0.20 & 15.84 & 0.15 & 17.27 & 0.16 \\
\hspace{2.7em}40\%  & \textbf{11.56}$^{*}$ & \textbf{0.31}$^{*}$ & \textbf{11.93}$^{*}$ & \textbf{0.31}$^{*}$ & \underline{17.87} & \textbf{0.07}$^{*}$ & \textbf{11.14}$^{*}$ & \textbf{0.29}$^{*}$ & \textbf{13.49} & \textbf{0.23}$^{*}$ & \textbf{14.36}$^{*}$ & \textbf{0.20}$^{*}$ \\
\hspace{2.7em}60\%  & 12.45 & 0.24 & 13.10 & 0.27 & 18.37 & 0.05 & 13.32 & 0.21 & 14.55 & 0.19 & 16.06 & 0.15 \\
\hspace{2.7em}80\%  & \underline{12.06} & \underline{0.28} & \underline{12.70} & \underline{0.30} & \textbf{17.45}$^{*}$ & \underline{0.06} & \underline{12.73} & \underline{0.22} & 14.11 & \underline{0.22} & \underline{15.48} & \underline{0.18} \\
\hspace{2.7em}100\% & 12.66 & 0.25 & 13.55 & 0.25 & 18.02 & 0.05 & 13.68 & 0.21 & 15.09 & 0.16 & 15.99 & 0.15 \\
\bottomrule
\end{tabular}%
}

\end{table*}

\noindent\textbf{Performance on challenging conditions and limited labeled data.} To validate our improvements under limited labeled multimodal data, we conduct a label-scarce cross-dataset study by varying the percentage of labeled subjects (randomly selected with five random seeds) used for supervision. The best RGB-only baseline (FactorizePhys) is trained with full-label supervision on the entire EquiPleth training set. In contrast, RPM-Distill uses all paired video--radar samples but only a subset (20--100\%) with labels for the joint objective $\mathcal{L}_{\text{sup}}+\lambda_{\text{distill}}\mathcal{L}_{\text{distill}}$; the remaining samples are trained with $\mathcal{L}_{\text{distill}}$ only. To avoid label leakage, the RF teacher is pretrained only on the labeled subset. Since we are unable to obtain other datasets that could provide RF and BVP signals, we skip the 0\% variant. We evaluate performance in PhysDrive across different illumination and motion conditions.

As reported in Table~\ref{tab:ssl}, FactorizePhys degrades substantially on challenging subsets, particularly at Night and under Talking, reflecting the low-SNR and motion-induced failure modes of video-only RPM. In contrast, RPM-Distill remains consistently more robust with limited supervision. Interestingly, the best performance is achieved at 40\% labeled data, while using more labels does not monotonically improve results. This is expected in our mixed objective setting: increasing the labeled fraction shifts optimization toward $\mathcal{L}_{\text{sup}}$ on the source-domain labeled subset, which can amplify domain-specific bias and reduce the regularizing benefit of leveraging a larger portion of paired data through $\mathcal{L}_{\text{distill}}$. Thus, 20--40\% provides a better balance between supervised anchoring and distillation-driven robustness, yielding the strongest gains across conditions.

\begin{table*}[t]
\centering
\scriptsize
\setlength{\tabcolsep}{0.8pt}
\renewcommand{\arraystretch}{0.92}
\scriptsize

% ---------- Left table ----------
\begin{minipage}[t]{0.49\linewidth}
\centering
\captionof{table}{Ablation study on PhysDrive and EquiPleth with cross-dataset setting.}
\label{tab:ablation_physdrive_equip}
\begin{tabular}{lccc ccc}
\toprule
\multirow{2}{*}{Method} & \multicolumn{3}{c}{PhysDrive} & \multicolumn{3}{c}{EquiPleth} \\
\cmidrule(lr){2-4}\cmidrule(lr){5-7}
& M.$\downarrow$ & R.$\downarrow$ & $r\uparrow$ & M.$\downarrow$ & R.$\downarrow$ & $r\uparrow$ \\
\midrule
w/o $\mathcal{L}_{\text{peak}}$   & 16.28 & 22.51 & 0.17 & 4.30 & 9.92 & 0.75 \\
w/o $\mathcal{L}_{\text{off}}$    & 16.10 & 22.37 & 0.16 & 3.40 & 8.29 & 0.82 \\
w/o $\mathcal{L}_{\text{shape}}$  & 15.95 & 22.29 & 0.16 & 3.13 & 7.93 & 0.83 \\
w/o weight  & 15.83 & 22.06 & 0.17 & 3.25 & 8.19 & 0.83 \\
w/o gate    & \underline{15.47} & \underline{21.66} & \underline{0.18} & 3.45 & 8.53 & 0.81 \\
w/o $\Phi(\cdot; \phi)$    & 15.68 & 21.89 & \underline{0.18} & \underline{2.78} & \underline{7.35} & \underline{0.85} \\
w/o bilevel  & 16.15 & 22.54 & 0.16 & 3.07 & 7.74 & 0.84 \\
Full        & \textbf{14.15} & \textbf{20.24} & \textbf{0.22} & \textbf{1.57} & \textbf{4.64} & \textbf{0.94} \\
\bottomrule
\end{tabular}
\end{minipage}
\hfill
% ---------- Right table ----------
\begin{minipage}[t]{0.49\linewidth}
\centering
\captionof{table}{Evaluation on MMPD and PURE. Models are trained on EquiPleth.}
\label{tab:cross_mmpd_pure}
\begin{tabular}{lccc ccc}
\toprule
\multirow{2}{*}{Method} & \multicolumn{3}{c}{MMPD} & \multicolumn{3}{c}{PURE} \\
\cmidrule(lr){2-4}\cmidrule(lr){5-7}
& M.$\downarrow$ & R.$\downarrow$ & $r\uparrow$ & M.$\downarrow$ & R.$\downarrow$ & $r\uparrow$ \\
\midrule
CHROM         & 13.63 & 18.75 & 0.08 & 9.79 & \underline{12.76} & 0.37 \\
POS           & 12.34 & 17.70 & \underline{0.17} & 9.82 & 13.44 & 0.34 \\
PhysNet       & 12.52 & 16.38 & 0.14 & 11.99 & 15.20 & 0.44 \\
PhysFormer    & 12.37 & 15.52 & 0.15 & 10.68 & 15.15 & 0.45 \\
FactorizePhys & \underline{12.15} & \underline{14.81} & \underline{0.17} & \underline{9.54} & 14.26 & \underline{0.51} \\
BeatFormer    & 12.28 & 15.34 & \underline{0.17} & 10.39 & 14.12 & 0.48 \\
\textbf{Ours}          & \textbf{11.60} & \textbf{12.34} & \textbf{0.19} & \textbf{8.28} & \textbf{10.33} & \textbf{0.55} \\
\bottomrule
\end{tabular}
\end{minipage}

\end{table*}

\begin{table}[t]
\centering
\caption{Robustness to training-time cross-modal misalignment under the EquiPleth$\rightarrow$PhysDrive protocol. The first column denotes the probability of applying such a shift to each training sample. }
\label{tab:shift}
\setlength{\tabcolsep}{4pt}
\renewcommand{\arraystretch}{0.9}
\scriptsize
\begin{tabular}{ccccccccc}
\toprule
\multirow{2}{*}{Shift Prob.} & \multicolumn{4}{c}{Ours} & \multicolumn{4}{c}{w/o gate} \\
\cmidrule(lr){2-5}\cmidrule(lr){6-9}
& STD$\downarrow$ & MAE$\downarrow$ & RMSE$\downarrow$ & $r\uparrow$ & STD$\downarrow$ & MAE$\downarrow$ & RMSE$\downarrow$ & $r\uparrow$ \\
\midrule
0   & \textbf{12.67}$^*$ & \textbf{14.15}$^*$ & \textbf{20.24}$^*$ & \textbf{0.22}$^*$ & 14.73 & 15.47 & 21.66 & 0.18 \\
0.2 & 14.50 & \underline{14.20} & \underline{20.30} & \textbf{0.22}$^*$ & 14.65 & 15.68 & 21.74 & 0.18 \\
0.4 & 14.69 & 14.75 & 20.81 & \underline{0.20} & 14.38 & 16.49 & 22.42 & 0.16 \\
0.6 & 14.65 & 14.77 & 20.80 & \underline{0.20} & \underline{14.30} & 17.04 & 23.50 & 0.15 \\
0.8 & 14.57 & 14.32 & 20.43 & \underline{0.20} & 14.69 & 17.88 & 24.91 & 0.13 \\
1.0 & 14.77 & 14.80 & 20.91 & 0.19 & 14.91 & 18.08 & 25.64 & 0.12 \\
\bottomrule
\end{tabular}
\end{table}

\noindent\textbf{Ablation study.} In the variant w/o weight, we disable the policy learner $
\Phi(\cdot; \phi)$'s output weights and use fixed, equal weights for the three distillation components. In the variant w/o gate, we remove the policy learner's output gate but keep its output weights. In the variant without $
\Phi(\cdot; \phi)$, we remove the entire policy learner and the associated bilevel meta-optimization.  The ``w/o bilevel'' variant keeps the same policy learner but optimizes it jointly with the student using $\mathcal{L}_{\text{train}}$. Table~\ref{tab:ablation_physdrive_equip} shows that each term in our physiology-structured objective improves robustness, where M./R./$r$ as MAE/RMSE/Pearson correlation. Dropping $\mathcal{L}_{\text{peak}}$ causes the largest degradation, especially on EquiPleth, indicating that peak mislocalization under challenging illumination is a dominant failure mode and that anchoring the student to the teacher's fundamental rhythm is crucial. Removing $\mathcal{L}_{\text{off}}$ or $\mathcal{L}_{\text{shape}}$ also consistently hurts performance, suggesting that suppressing the off-peak noise floor and constraining spectral morphology provide complementary benefits beyond peak alignment. Besides, using fixed, equal component weights or removing the gate degrades performance, implying that the optimal spectral focus is sample-dependent and indiscriminate supervision can cause negative transfer. Finally, removing $\Phi(\cdot;\phi)$ leads to additional drops, showing that policy learning is key for strong generalization.

\noindent\textbf{Case study.} To assess whether the distillation improves the student beyond the multimodal benchmarks, we further perform a case study where the models are trained on a limited-size dataset (EquiPleth) is directly evaluated on RGB-only datasets. As shown in Table~\ref{tab:cross_mmpd_pure}, our distilled student consistently outperforms all RGB baselines on both MMPD and PURE, achieving the lowest MAE/RMSE and the highest r. This suggests that the proposed method transfers a modality-agnostic rhythm prior that remains beneficial even when radar is absent at both training and deployment on the target domain. In particular, the gains on MMPD, which contains diverse activities and more severe motion, indicate that RPM-Distill can leverage RF-guided training to produce a stronger video-only model that generalizes to unseen RGB-only domains.

\noindent\textbf{Robust to unsynchronized multimodal training data.} We evaluate the robustness of the learned gate by explicitly corrupting cross-modal alignment in the well-synchronized EquiPleth training set. For a portion of training pairs, we shift the RF window forward by 30 steps (1s) while keeping the RGB clip unchanged, and vary the shift probability before testing on PhysDrive. As shown in Table~\ref{tab:shift}, the ungated variant degrades steadily as the shift probability increases, indicating accumulated negative transfer from mismatched RF supervision. In contrast, the full model remains relatively stable and consistently outperforms, even when all training pairs are shifted. This suggests that the learned gate can down-weight unreliable student--teacher spectral relations and is effective beyond ideally synchronized training conditions.

\subsection{Visualization}

\begin{figure*}[t]
\begin{center}
\includegraphics[scale=0.29]{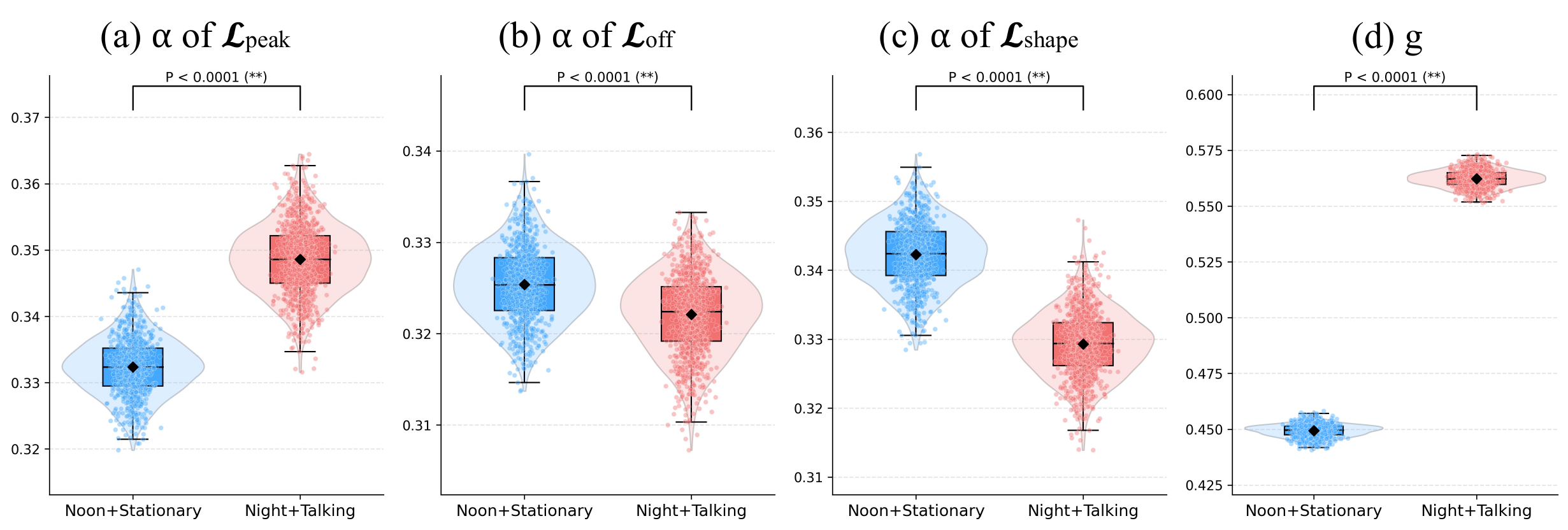}\\
\end{center}
\vspace{-5mm}
\caption{A visualization of component weights \(\boldsymbol{\alpha}\) and distillation gate \(g\) distributional changes when training in different PhysDrive scenarios. We measure the difference by the paired t-test.}\label{distribution}

\end{figure*}

\begin{figure*}[t]
\begin{center}
\includegraphics[scale=0.06]{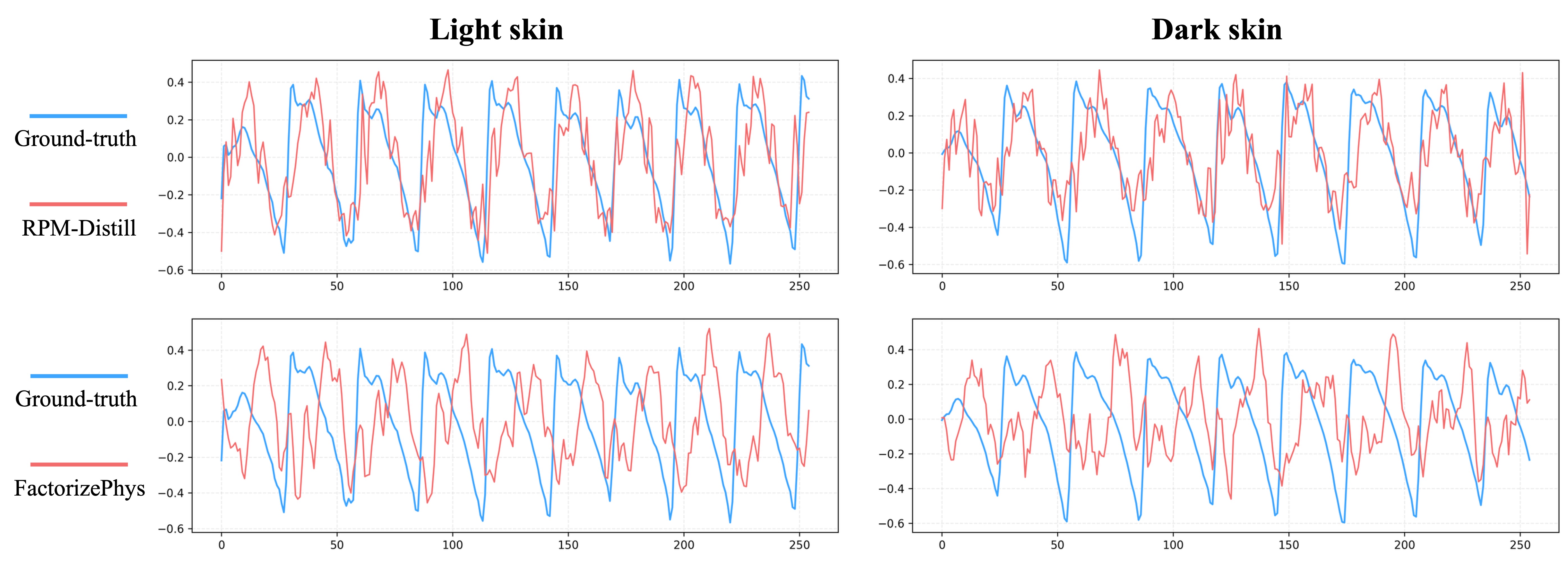}\\
\end{center}
\vspace{-5mm}
\caption{Visualization of pulse waveform examples generated by RPM-Distill and FactorizePhys in light and dark skin tone groups of EquiPleth.}\label{signal}

\end{figure*}

We assess the effectiveness of our dynamic distillation policy by visualizing component weights and gate values when training on different scenarios. As shown in Fig.~\ref{distribution}, the learned policy shows systematic distributional shifts between an easier subset (Noon+Stationary) and a harder subset (Night+Talking) on PhysDrive, with all changes statistically significant under paired $t$-tests. In the harder scenario, the policy assigns a larger gate $g$, meaning student training relies more on RF guidance when optical RPM becomes unreliable due to low illumination and stronger motion. The policy also increases the weight of the peak-related component while down-weighting the others. This matches our design motivation: under challenging conditions, video RPM primarily fails via peak suppression/shift and spurious maxima, so emphasizing rhythm anchoring around the fundamental frequency is more effective than time-domain matching.

Besides, Fig.~\ref{signal} provides additional qualitative evidence from PhysDrive to EquiPleth by comparing representative pulse waveforms from RPM-Distill and an RGB baseline without distillation, across different skin tones. In the light-skin group, both methods recover a periodic structure, but RPM-Distill yields cleaner, more temporally consistent cycles with fewer jagged fluctuations. In the dark-skin group, where optical SNR is lower, RPM-Distill maintains a clearer quasi-periodic pattern with more coherent peak timing, consistent with metric gains. It suggests that our method adapts the distillation focus under difficulty shifts, and that RF-guided spectral distillation improves robustness to skin-tone-induced domain gaps while preserving video-only inference.

% \noindent\textbf{Limitations and future work.} 
\section{Discussion}
Our formulation focuses on band-limited spectral distillation and HR-centric evaluation; extending it to richer waveform-fidelity objectives and additional vital signs (e.g., RR or multi-task settings) is an important direction. RPM-Distill also assumes synchronized video–RF pairs during training and uses a small labeled validation split for bilevel policy learning, which may limit applicability under severe desynchronization, missing radar segments, or extremely scarce labels. Due to the limited availability of public RF datasets with consistent ground truth, we cannot yet fully validate unsupervised cross-dataset distillation, where an RF teacher pretrained on one dataset transfers to another without target labels. Moreover, our spectral interface presumes a stable periodic rhythm within a predefined physiological band; atypical rhythms, strong nonstationarity, or camera-dependent sampling differences may challenge this assumption. Finally, emerging multimodal foundation models may offer alternative routes for cross-modal alignment, and a careful comparison with physiology-structured distillation remains open.

Beyond accuracy, RPM-Distill offers a practical perspective on how to operationalize complementarity under modality mismatch. For RPM, RGB, and RF are not commensurate in the time domain and may exhibit modality-specific distortions; our results indicate that a more reliable transfer interface is the band-limited spectral rhythm, where heterogeneous sensors share a common periodic substrate. Decomposing distillation into three spectral components is sufficient to target dominant failure modes, while the learned policy highlights that both how much and what to distill are sample-dependent. This design also suggests a transferable route for heterogeneous-sensing tasks beyond RPM. When different modalities capture the same latent process through distinct physical mechanisms, RPM-Distill can be instantiated by defining the task-relevant shared structure, decomposing it into meaningful distillation components, and learning sample-adaptive reliability weights for cross-modal transfer.

\section{Conclusion}

In this paper, we presented \textbf{RPM-Distill}, a physiology-guided, adaptive cross-modal distillation framework that leverages RF radar in training to improve video-based RPM while preserving video-only inference. Unlike feature-level transfer or indiscriminate output matching, RPM-Distill distills shared, physiology-structured spectral evidence and uses a data-driven policy to gate and reweight supervision, reducing negative transfer under RF corruption or imperfect synchronization. The bilevel meta-objective further anchors policy learning to generalizable measurement \cite{wang2023hierarchical,wang2024generalizable}. Extensive tests show consistent improvements.

% \noindent\textbf{Limitations and future work.} Our formulation focuses on band-limited spectral distillation and HR-centric evaluation; extending it to richer waveform-fidelity objectives and additional vital signs (e.g., RR or multi-task settings) is an important direction. RPM-Distill also assumes synchronized video–RF pairs during training and uses a small labeled validation split for bilevel policy learning, which may limit applicability under severe desynchronization, missing radar segments, or extremely scarce labels. Due to the limited availability of public RF datasets with consistent ground truth, we cannot yet fully validate unsupervised cross-dataset distillation, where an RF teacher pretrained on one dataset transfers to another without target labels. Moreover, our spectral interface presumes a stable periodic rhythm within a predefined physiological band; atypical rhythms, strong nonstationarity, or camera-dependent sampling differences may challenge this assumption. Finally, emerging multimodal foundation models may offer alternative routes for cross-modal alignment, and a careful comparison with physiology-structured distillation remains open.

\section*{Acknowledgements}
We gratefully acknowledge financial support from the Natural Sciences and Engineering Research Council of Canada (NSERC) Discovery Grant program, the Canada Research Coordination Committee’s New Frontiers in Research Fund (CRCC NFRF) Exploration, and the Canadian Foundation for Innovation (CFI) John R. Evans Leaders Fund (JELF). 

\bibliographystyle{splncs04}
\bibliography{main}
\end{document}